\documentclass[journal]{IEEEtran}
\usepackage{graphicx}
\usepackage{cite}
\usepackage{booktabs}
\usepackage{comment}
\usepackage{bm}
\usepackage{multirow} 
\usepackage{amsmath,amssymb}
\ifCLASSINFOpdf
\else
\fi
\begin{document}
%
\title{A Random-patch based Defense Strategy Against Physical Attacks for Face Recognition Systems}
%
%
%

\author{{JiaHao~Xie,~Ye~Luo,~\IEEEmembership{Member,~IEEE,}~Jianwei~Lu}
\thanks{J. Xie, Y. Luo and J. Lu are with the department of Software Engineering, Tongji University, Shanghai 201804, China (email: jh\_xie@tongji.edu.cn; yeluo@tongji.edu.cn; jwlu33@tongji.edu.cn)}
}

%
%

\markboth{Journal of \LaTeX\ Class Files,~Vol.~14, No.~8, October~2020}%
{Shell \MakeLowercase{\textit{et al.}}: Bare Demo of IEEEtran.cls for IEEE Journals}
%



\maketitle

\begin{abstract}
The physical attack has been regarded as a kind of threat against real-world computer vision systems. Still, many existing defense methods are only useful for small perturbations attacks and can't detect physical attacks effectively. In this paper, we propose a random-patch based defense strategy to robustly detect physical attacks for Face Recognition System (FRS). Different from mainstream defense methods which focus on building complex deep neural networks (DNN) to achieve high recognition rate on attacks, we introduce a patch based defense strategy to a standard DNN aiming to obtain robust detection models. Extensive experimental results on the employed datasets show the superiority of the proposed defense method on detecting white-box attacks and adaptive attacks which attack both FRS and the defense method. Additionally, due to the simpleness yet robustness of our method, it can be easily applied to the real world face recognition system and extended to other defense methods to boost the detection performance.

\end{abstract}


%
\IEEEpeerreviewmaketitle

\section{Introduction}
%
%
%
%
\IEEEPARstart{D}{eep} neural networks (DNNs) achieve exceeding success during recent years and even outperform humans in some domains such as hand-written digits recognition and face recognition. Although DNNs work well in many tasks, crafted patterns designed by the adversarial attacks  can still fool DNNs easily.

Face Recognition System (FRS) deployed with DNNs are facing a considerable risk of being attacked. Fortunately, FRS takes pictures with a built-in camera before executing an algorithm to detect the threat. In this circumstance, attackers are not allowed to change digital images, and generating small perturbations attacks by directly manipulating digital images are impossible. However, in the real world, defending FRS from being attacked is always a big issue. Unlike other recognition tasks for natural images, FRS has so much close relationship to the personal and the property safety, and the most severe threat to FRS is the physical attacks. Different from general adversarial attacks, physical attack releases the constraint such that any perturbation changing pixel values within 0-255 is allowed. Moreover, the pattern produced by physical attacks can be transferred to the models in FRS even after transforming images by the camera. It exposes DNN based FRS to a severe threat as long as physical attacks exists. Hence, it's crucial to design a reliable defense model against physical attacks in the real world.

\begin{figure}
\centering
\includegraphics[height=3.8cm]{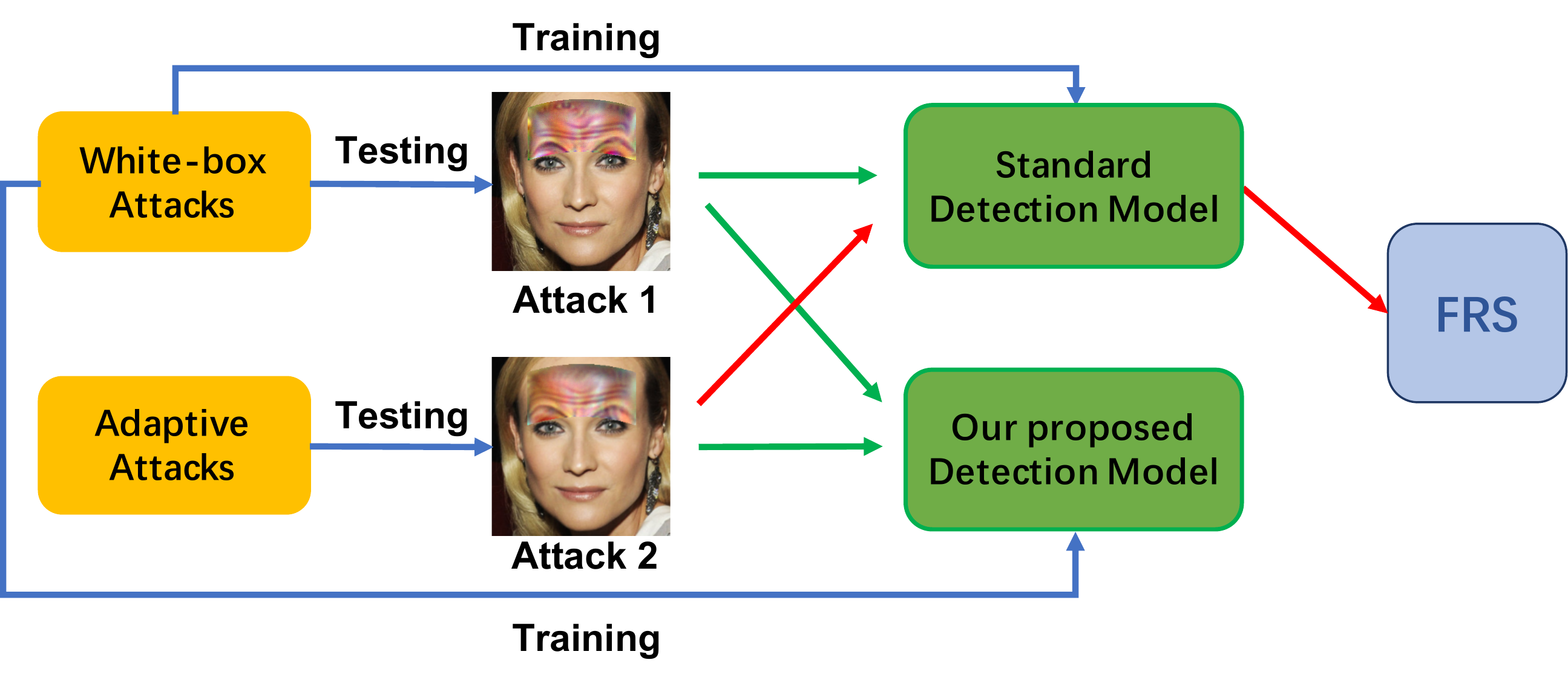}
\caption{The main idea of the proposed physical attack defense model for Face Recognition System (FRS). The standard detection model is trained with normal images and white-box attacks but can't resist adaptive attacks like defense-strategy-leaked attacks and defense-model-leaked attacks. While our proposed detection model is more robust and can defend FRS from being attacked by white-box attacks and various adaptive attacks. }
\label{fig:intro}
\end{figure}

Some defense methods\cite{tao2018attacks,wu2020defending} are also proposed to help DNNs based FRS from being attacked by physical attacks. At first, defenses methods with complex models\cite{wu2020defending} are not acceptable in the real world, considering that it's very expensive to retrain a model in FRS. A lightweight model with a high detection accuracy rate is expected. Secondly, many defense methods focus on resolving imperceptible perturbations that distribute on the whole image. However, only the limitation that these methods are vulnerable to attacks with bigger perturbations is noticed \cite{carlini2017towards,madry2017towards}, and the way to detect attacks globally from a whole image is rarely being questioned. In other words, few research put their focus on the \textbf{defense strategy} (e.g. detecting from the whole image or the local image patch). Thirdly, some methods are proposed to detect physical attacks but the accuracy rates of them are still far from being applied to the real-world system, such as \cite{tao2018attacks} achieves 85\% detection accuracy and obtains 9.91\% false positive rate against the glasses-attack, and the methods in \cite{wu2020defending} can classify identity which attacked by glasses-attack\cite{sharif2016accessorize} with a accuracy about 85\%. However, they are far from satisfying the real-world requirements of FRS. Moreover, the method in \cite{tao2018attacks,wu2020defending} doesn't verify their defense method against the \textbf{defense-strategy-leaked attack}, in which the attacker has no access to the model parameters but knows the defense strategy. And the defense-strategy-leaked attack is most likely to happen in the real world in that the attackers could probably guess the defense strategy, but it is difficult for them to know the parameters of the model exactly. At last, there are also some challenging attacks which can access both the defense strategy and the detailed parameters of the defense model named as \textbf{defense-model-leaked attack}, which does exist but is barely used to verify the existing defense methods.

Except for the aforementioned problems existing in the previous methods, some valuable phenomenons are also noticed in detecting the attacks in a face image. (1) Splitting an image into small patches can destruct the structure of an attack for the whole image. (2) Once an attack is detected on an image patch, the whole image is considered to be attacked. (3) Attacking many patches simultaneously is exponentially difficult compared to attacking one whole image. In other words, dividing the image into patches can destruct the strategy of the attack and make the attack more difficult than attack one whole image. Moreover, previous methods are also proposed to utilize the image patches to detect face spoofing \cite{atoum2017face,de2018learning}. But their purpose is to utilize local features, while we aim to divide a whole image into small patches to detect physical attacks. To the best of our knowledge, few work is proposed to use it for defending FRS against physical attacks. 

In order to address the aforementioned problems, we propose a random-patch based defense model which defends FRS against various physical attacks by inspecting the inputs before feeding them to the face recognition system. The main idea is shown in Fig.1. Specifically, we split the images into image patches either evenly or randomly, and then train the detection model to filter out suspicious before feeding the face image to a FRS. We train the detection model only once purely by the normal images and the white-box attacks which are generated by attacking the Arcface model only \cite{sharif2016accessorize,komkov2019advhat}. Once the model is trained, it can be used to detect various attacks including the white-box attacks, the defense-strategy-leaked attacks, and the defense-model-leaked attacks. In order to obtain the defense-strategy-leaked attacks and the defense-model-leaked attacks, a multi-task attack model is particularly designed by attacking a face model and a defense model simultaneously. Given a defense model to be verified, the multi-task attack model can accordingly change its employed defense strategy or defense model. We name the attacks generated by multi-task attack model as the \textbf{adaptive attacks}. In other words, if the defense strategy of the multi-task attack model and the verified defense model is the same, we call it as the defense-strategy-leaked attack. If both the defense strategy and the defense model of the two models are the same, we call it as the defense-model-leaked attack. For the other two cases (i.e. none of them are leaked, and only the model is leaked), we didn't consider in this paper due to its rareness in real world and simpleness to group them to the above three attack categories. At last, we perform extensive experiments on the public VGGFace dataset \cite{parkhi2015deep} and verify the effectiveness of the proposed defense model against the attacks including but not limited to the white-box attacks, the defense-strategy-leaked attacks, and the defense-model-leaked attacks. And our best defense model achieves 100\% detection accuracy against the white-box attack.

The main strengths of the proposed defense model lie onto the following three aspects. (1) Compared to some complex defense methods \cite{tao2018attacks,wu2020defending}, our proposed defense model is a lightweight model in that it is based on a single DNN and only needed to be trained once. (2) Compared to the defense strategy by detecting attacks on the whole image, the image patch based defense strategy is robust and difficult to be attacked. Also, the use of randomness further improves the robustness of the proposed defense model. (3) We achieved superior detection accuracy against various attack on the public VGGFace dataset \cite{parkhi2015deep}, which demonstrates the feasibility of our method to be applied into the real world.

Our contributions can be concluded in three folds:

\begin{itemize}[\IEEEsetlabelwidth{Z}]
\item We propose a random-batch based defense model to prevent physical attacks against the face recognition system, which is simple yet robust and flexible to be used in the real world.
\item Our method achieves nearly 100\% accuracy on the detection of the white-box hat-attack and glasses-attack.
\item We design a multi-task attack model to generate adaptive attacks including the defense-strategy-leaked attacks and the defense-model-leaked attacks, and they are further used to verify the effectiveness and robustness of our proposed defense method compared to the standard detection method.
\end{itemize}

\section{Related Work}

In this part, we first introduce recent work about adversarial attacks, then we explore defense methods to construct a robust model.

\subsection{Attack Methods}
\subsubsection{Adversarial Attack}
Adversarial attacks are the phenomenon in which machine learning models can be tricked into making false predictions by slightly modifying the input. 
In \cite{szegedy2013intriguing}, they first find that small imperceptible perturbations searched by L-BFGS can fool models. Shortly afterward, the authors in \cite{goodfellow2014explaining}
propose the fast gradient sign method (FGSM) to find an adversarial example efficiently. This method needs just one step to compute an adversarial attack within ${\epsilon}$ perturbation by $L_{\infty}$ constraints. FGSM is the first algorithm that employs one-step gradient descent in optimization for the model attack, and this inspired many other methods which adopted various optimization tricks for better attack performance. In \cite{kurakin2016adversarial}, they find that computing perturbations iteratively can generate a stronger attack than FGSM. The authors in \cite{dong2018boosting} enhance the transferability of adversarial attacks by adding momentum to overcome the risk of overfitting and win the top in NIPS 2017 Non-targeted Adversarial Attack
and Targeted Adversarial Attack competitions. C\&W \cite{carlini2017towards} treat the adversarial attack as an optimization problem and design a loss function to get incorrect prediction with a relatively small $L_2$ distance. PGD\cite{madry2017towards} is the state of the art algorithm to produce first-order adversarial attacks. The key ideas of how PGD is applied to are:
(1) every step computes the gradient respect to an image after adding a small random noise to the original image.
(2) divide ${\epsilon}$ into $m$ parts and apply m times FGSM algorithm to achieve the final result. Thinking that exchanging the features closest to the classification decision boundary is a much easy way to construct an attack, Deepfool\cite{moosavi2016deepfool} implements their idea by approximating the no-linear problems via many small linear ones at every step.

\subsubsection{Physical Attack}
Instead of restricting the distance (such as $L_2$ distance) between the attack and the original image, the physical attack usually occurs within a certain region in which the attack can reach\cite{sharif2016accessorize,brown2017adversarial,komkov2019advhat}. The method in \cite{sharif2016accessorize} prints a pattern attached to the glasses and fool the FRS successfully. The authors in \cite{brown2017adversarial} proposed to use a circular patch to construct a universal, robust attack in the real world. Attaching a crafted pattern on the forehead also can fool the state of the art Face ID model \cite{komkov2019advhat}.

\subsection{Defense Methods}
\subsubsection{Adversarial Training }
Adversarial training is a simple and effective way to construct robust defense models \cite{goodfellow2014explaining,carlini2017towards,madry2017towards}. The main idea of adversarial training is to train a model using data that contains both original images and their adversarial attacks. But the model trained by adversarial training is only robust to attacks which appear in the training set, and a stronger attack can break the defense easily. In addition, a recent work\cite{wu2020defending} focus on defending physical attack proposed a method named DOA, which does adversarial training by using attacks generate by ROA and trains a model achieved about 85\% accuracy in a small dataset contains ten identities.

\subsubsection{Recovering natural distribution}
Methods of this kind treat adversarial attacks as small noise and try to remove those noise in the pre-processing. Defense-GAN \cite{samangouei2018defense} uses GAN \cite{goodfellow2014generative} to train a generative model to simulate the distribution of unperturbed images and then uses the generative model to generate the images similar to the input images by applying random times optimization. \cite{sun2019adversarial} projects images into a quasi-natural image space through adding a spare transformation layer using convolutional sparse coding\cite{bristow2013fast,choudhury2017consensus}. \cite{guo2017countering} proposes to use image quilting\cite{efros2001image} to substitute unknown inputs with natural image patches from the employed dataset.

\subsubsection{Feature squeezing}
Feature squeezing makes the attack difficult by reducing the space that an attacker could manipulate. Such as in \cite{dziugaite2016study}, the authors found that JPG compression can boost the classification accuracy of adversaries. Meanwhile, the authors in \cite{xu2017feature} propose to squeeze features space by reducing the color depth or blurring it via the spatial smoothing. 

\subsubsection{Extracting robust features}
The robust feature is the features that are insensitive to small perturbations. In \cite{gao2017deepcloak}, a mask is used to filter sensitive features after a DNN extractor. In \cite{tao2018attacks}, the authors consider the features related to five sense organs are crucial to the FRS and propose to construct another model to perform auxiliary inference by strengthening witness features and weakening non-witness features.

\section{The Proposed Method}
Since our proposed method is the physical attack defense model, and in this Section, we first introduce two types of physical attacks and their corresponding implementation methods. Then, we introduce our designed multi-task attack model, which can generate adaptive attacks on both the face model and the detection model simultaneously. At last, we introduce the proposed random-patch based defense strategy, which can be applied to a broad of DNNs based defense models and protect the FRS from both the white-box attacks and the adaptive attacks.

\begin{figure}
    \begin{minipage}[t]{0.48\linewidth}
        \centering
        \includegraphics[width=1.5in]{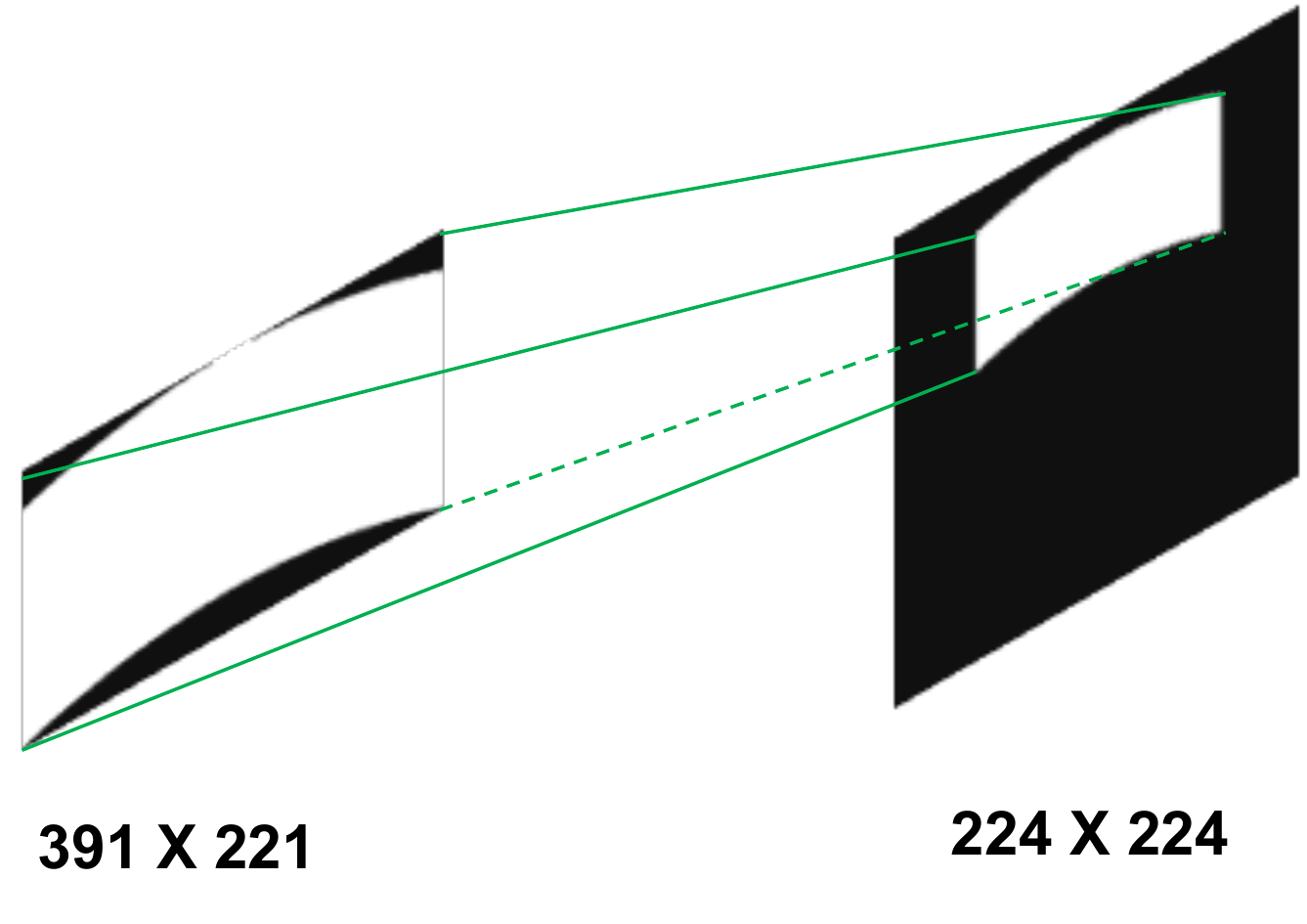}
    \end{minipage}
    \begin{minipage}[t]{0.48\linewidth}
        \centering
        \includegraphics[width=1.5in]{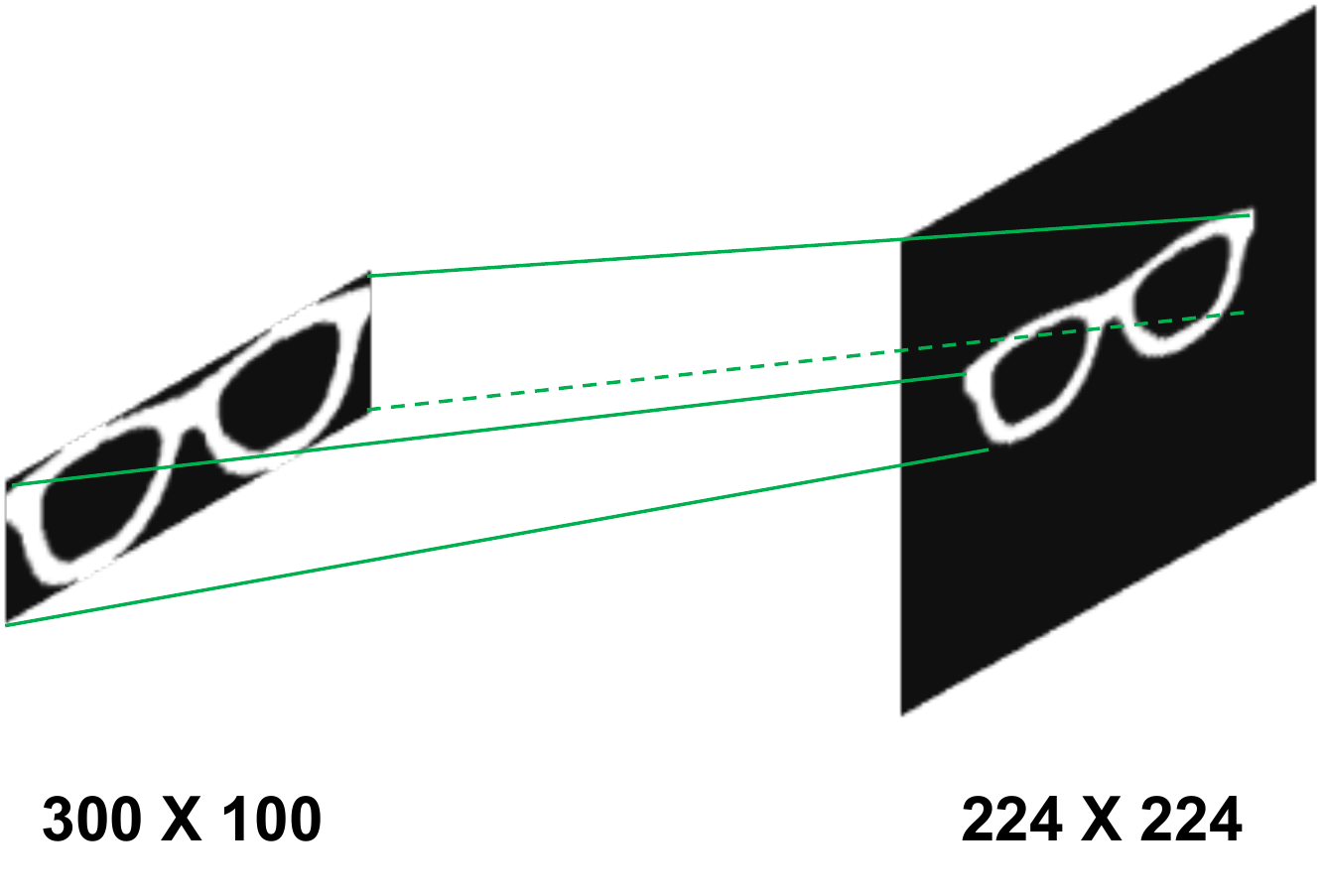}
    \end{minipage}
    \caption{Illustration of different masks of attacking regions which are projected by the Spatial Transformer Network (STN). Left is the mask of hat-attack and right is a mask of glasses-attack.}
    \label{fig.1}
\end{figure}

\subsection{Physical Attack Generation}
For the first physical attack, we reproduce the hat-attack \cite{komkov2019advhat} but modify a little to adapt to our experiment. We replace the project process in \cite{komkov2019advhat} with a certain mask sampled from $391\times221$ to $224\times224$ by a predefined STN (Spatial transformer network)\cite{jaderberg2015spatial} as shown in Fig.\ref{fig.1} left. 


Similar to the hat-attack generation, we generate the second physical attack named the glasses-attack \cite{sharif2016accessorize}. The same training process is adopted as in hat-attack \cite{komkov2019advhat}, and the attack mask of the glasses region is sampled from $300\times100$ to $224\times224$ by another predefined STN, as shown in Fig. \ref{fig.1} right.

After the generation of the mask of the attack region, we can obtain the raw attack by fusing the mask to the original image. Generally, there are two ways to implement the attack action: the dodging attack and the impersonation attack. In a dodging attack, the aim is to decrease the similarity between the attack and the original image. Meanwhile, the impersonation attack aims to increase the similarity between the attack and the target image. Here the target image is another person that the attack wants to impersonate.

\begin{figure*}[!t]
\centering
\includegraphics[width=14cm]{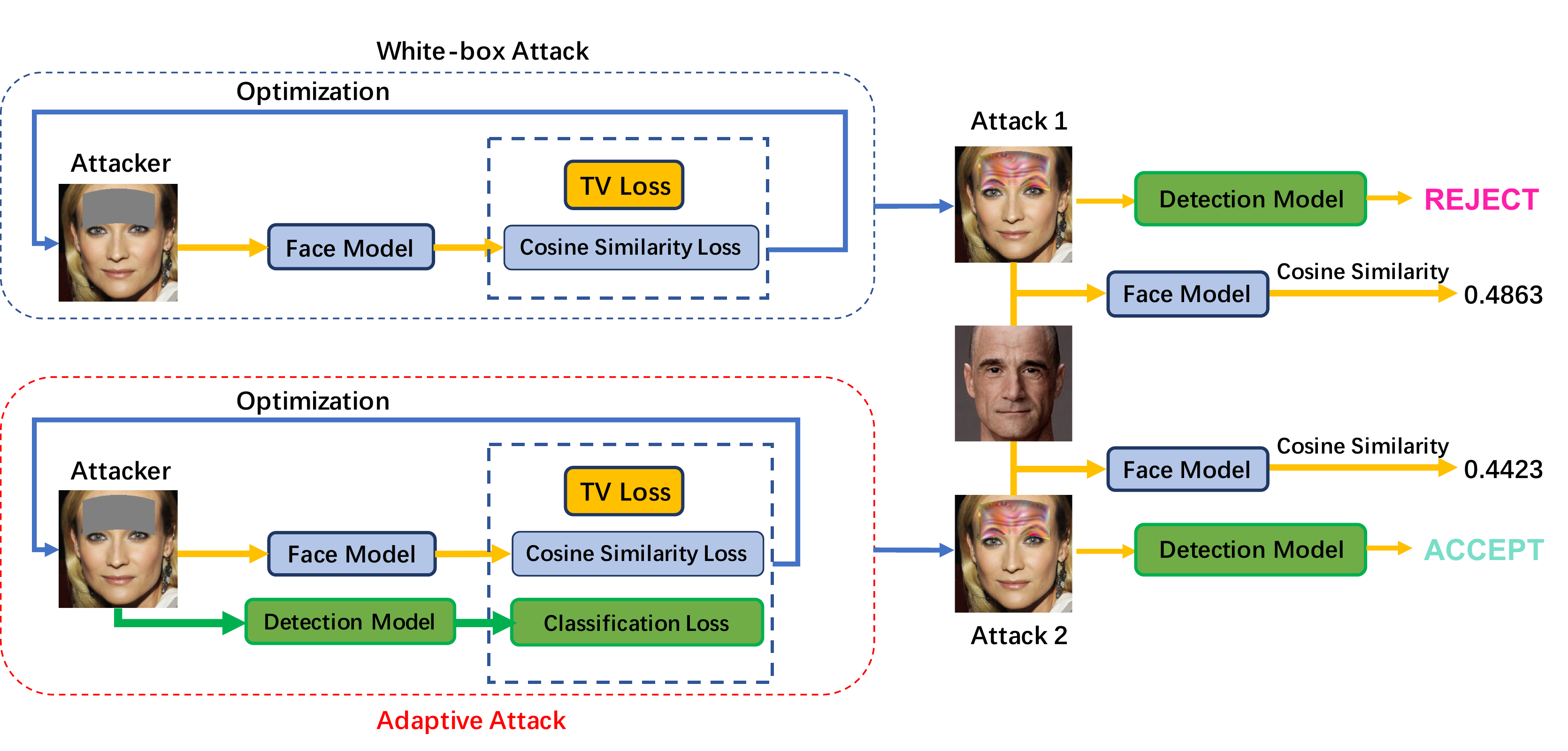}
\caption{Illustration of the procedures to generate the white-box attack and the adaptive attack. Attack1 is generated by the white-box attack which can be rejected by a detection model such as a standard DNN easily, and the cosine similarity between Attack1 and the Target is 0.4863. While Attack2 belonging to the adaptive attack which is generated by our multi-task attack model can attack the face model and fool the standard detection model simultaneously.}
\label{fig:attack-process}
\end{figure*}

\subsection{Multi-task Attack Model}
While DNN based method can detect physical attacks easily, defense strategies with a standard DNN are still unsafe. The adaptive attacks can fool the face model and the detection model simultaneously. Take the impersonation hat-attack as an example, which is shown in Fig.\ref{fig:attack-process}.  Given the attack and the target face image, a common attack (i.e., Attack1 in Fig. \ref{fig:attack-process} ) generated by the loss function with the similarity and the TV loss can be successfully detected by an existing attack detection method. However, an adaptive attack (i.e., Attack2 in Fig.\ref{fig:attack-process}) can not only attack a face model but also fool an existing attack detection model. 

To model attacks like Attack2, we design a multi-task attack model to generate attacks that can attack a face model and a detection model simultaneously. The loss function of ours is composed of three items: the cosine similarity loss to measure the similarity between the attack and the original image, the TV loss to smooth pixel values in the generated attack, and the classification loss to implement the attack to the detection model.

Thus, in this paper, the proposed loss function is:
\begin{align}
     L = L_{sim} + \alpha L_{cls} + \beta L_{tv},
\end{align}%
where $L_{sim}$ denotes the loss of similarity between the attack and the target. $L_{cls}$ is used to attack a face model. $\alpha$ is used to control the attack strength of $L_{cls}$. $L_{tv}$ is the TV loss, and $\beta$ is a weight for the TV loss.

 $L_{sim}$ is defined as:
\begin{equation}
L_{sim}=\left\{
\begin{aligned}
& {Sim}_{(attack, original)}     & dodging \\
& -{Sim}_{(attack, target)}    & impersonation
\end{aligned}
\right.
\end{equation}
where $Sim(a, b)$ denotes the cosine similarity between image a's feature and image b's feature extracted by the face model (In this paper ArcFace model is used.), the formula is:
\begin{equation}
    Sim(a, b) = \frac{\boldsymbol {e_a \cdot e_b}}{{\Vert\boldsymbol{e_a}\Vert}_2 {\Vert\boldsymbol{e_b}\Vert}_2},
\end{equation}

\noindent where $\bm{e_a}$ and $\bm{e_b}$ are the features extracted by a face model for the image a and the image b, respectively. $\bm{\cdot}$ means inner product, and $\bm{\Vert\Vert}_2$ is $L_2$ norm of the vector.

$L_{cla}$ aims to fool the detection model, and it equals to zero when there is no attack to a detection model. Otherwise, it is equal to:
\begin{equation}
    L_{cls} = -\frac{1}{N} \sum_{i=1}^N (p_i^0 \log{q_i^1} + p_i^1 \log{q_i^0}),
\end{equation}%
where $q_i^c$ is the softmax output of the detection model, which shows the probability that the $i_{th}$ patch in the face image is classified as label $c$. $p_i^c$ is the true probability of labeling as $c$ for the $i_{th}$ patch in the face image. $N$ is the total number of image patches and $C$ is the number of class labels. This loss forces the attack to avoid being detected by the detection model.


\begin{figure}
    \begin{minipage}[t]{0.48\linewidth}
        \centering
        \includegraphics[width=0.9in]{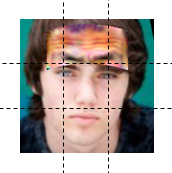}
        \label{fig.2.a}
    \end{minipage}
    \begin{minipage}[t]{0.48\linewidth}
        \centering
        \includegraphics[width=0.9in]{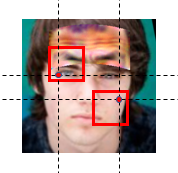}
        \label{fig.2.b}
    \end{minipage}
    \caption{Illustration of two ways to split an image into patches. The left is the way to evenly split the image into nine patches, while the right is to randomly split the images by two randomly selected points.}
    \label{fig.2}
\end{figure}

\subsection{Random-patch based Defense Strategy}
Since DNN is vulnerable to adversarial attacks, attackers can attack the face model and the detection model simultaneously. We propose a random-patch based defense model to detect attacks, such that it is more robust than standard DNNs.

\subsubsection{Why image patches? }
There are multiple advantages to employ image patches in a defense model. At first, the attacked region in every patch is big enough to be detected, and an attack on the whole image also loses its aggressivity when the image is cropped into pieces. Last but not least, previous attacks need to attack one image and only fool the detection model once. But in our method, the attacks need to attack all patches successfully. That's the main reason that our defense model is more robust than standard DNNs.  Considering all these reasons, we evenly divide the image into nine patches, as shown in Fig. \ref{fig.2} left, and an image patch based defense strategy is adopted in our proposed defense method.

\subsubsection{Why random patches? } 
Evenly splitting the image into patches can against some physical attacks. However, attackers may also attack a patch based defense model, especially when our defense methods are leaked. Motivated by \cite{guo2017countering} that defenses with random strategies are more robust than a specific transform. We propose a random-patch based defense method, which replaces the evenly divided image patches by the ones with the random sizes. Specifically, two points $(x1,y1)$ and $(x2,y2)$ are first randomly selected, where $x1 \in [W/3-W/6, W/3+W/6]$, $x2 \in (2W/3-W/6, 2W/3+W/6]$, $y1 \in [H/3-H/6, H/3+H/6]$, and $y2 \in (2H/3-H/6, 2H/3+H/6]$. Here $W$ and $H$ correspond to the width and the height of the image, respectively. According to the two selected points, we can randomly split the image into nine patches with different sizes, as shown in Fig. \ref{fig.2} right.
Later in the experiment, we will show that our random-patch based defense method is robust to while-box attacks and even valid to adaptive attacks.

\begin{figure*}[!t]
\centering
\includegraphics[width=14cm]{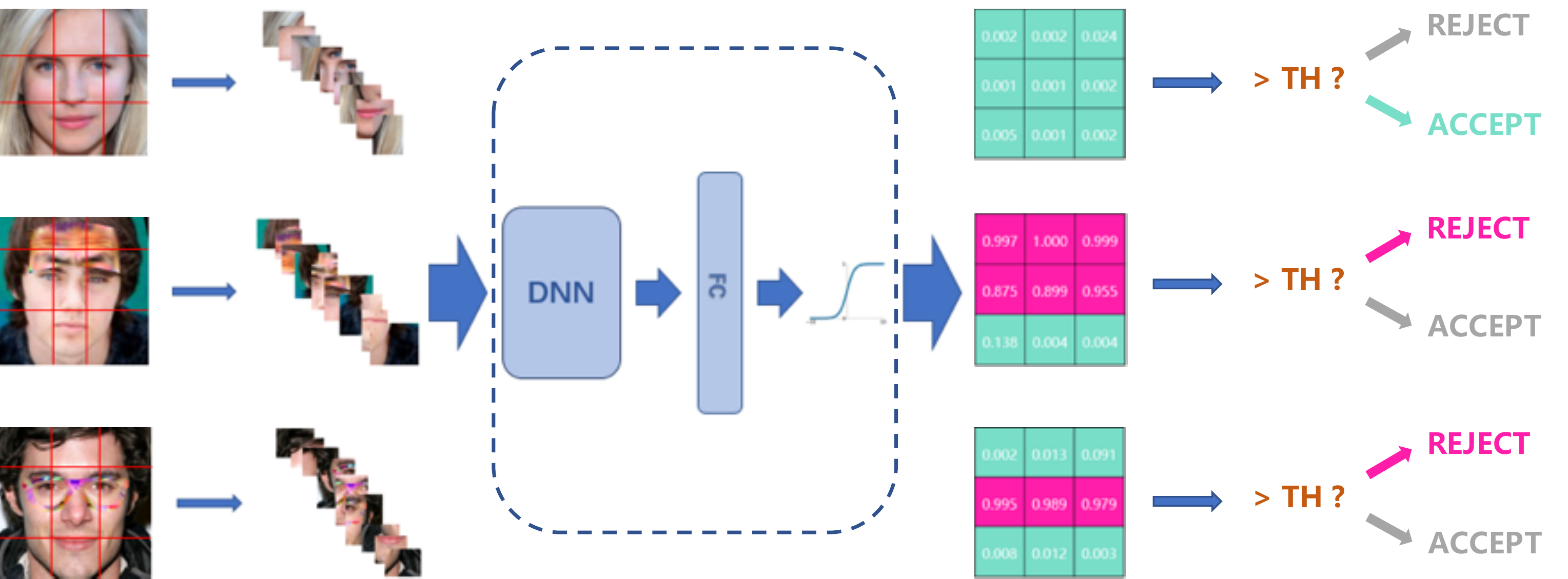}
\caption{The whole procedure of our proposed random-patch based defense method. Three examples are shown here. Each of the face images is first cropped into nine patches. And then, for each image patch, we use a DNN model with a softmax function to compute how probably it is being attacked. A face image with  no less than $TH$ attacked patches will be rejected by the proposed defense method.} 
\label{fig:denfense-pipline}
\end{figure*}

\subsubsection{How to train and inference? } Training our random-patch based defense model is totally different from training a standard DNN. Specifically, for every step, we choose a batch of images from the training set randomly, then we divide every image into nine patches (either evenly or randomly based on the model). Patches of a normal image labeled as 0, while the label of a patch from an attack is decided according to the number of pixels falling into the attacked region. When the number of pixels falling into the attacked region is larger than a predefined threshold $th$ (e.g., $th=800$, and it is about 1.6\% of the original input image area $224\times224$.), we label the patch as 1. Otherwise, it is directly labeled with the ratio $\frac{Num_{pixels}}{th}$.

For every normal image, we randomly choose six out of nine patches, and for every attack image, we choose the top six patches with the large label values. At last, after resizing all these image patches into a fixed size ($224\times224$), we use those image patches to train the DNN model by the batch gradient descent. 

In inference, given a test face image, we first randomly divided it into nine patches. And then, for each image patch, we use the trained DNN model with a softmax function to inference how probably it is being attacked. After thresholding the probability, we can judge whether the patch is being attacked or not. The pipeline is shown in Fig. \ref{fig:denfense-pipline}. A face image with no less than $TH$ attacked patches will be rejected by the defense method. Here, we set $TH=1$ such that any face image contains one attacked patch will be rejected by the proposed defense method. 

\section{Experiments and Analysis}
In this section, we conduct the experiments to verify the effectiveness of our proposed random-patch based defense strategy. At first, we introduce the dataset, the experimental settings, and the effectiveness of attacks generated by the proposed multi-task attack model. Then, we conduct the experiments for our proposed defense method against the white-box attack, the defense-strategy-leaked attack and the defense-model-leaked attacks. At last, in order to further validate the robustness of the proposed defense method, we verify it on attacks generated with various face models, on different mask shapes and even different datasets.
\subsection{Dataset}
Experiments are performed on VGGface dataset\cite{parkhi2015deep}, which contains 2622 identities. Each identity has 1000 images, and totally it has 2.6M images. This dataset is too big to verify our method, so we choose one picture from one identity. For every picture, we crop and align the face region by MTCNN\cite{zhang2016joint}, then separate images into the training set, the validation set and the testing set with 7:1:2 proportion. Finally, we have 1834 images in the training set, 262 images in the validation set, and 526 images in the testing set.

\subsection{Experimental Settings}
\subsubsection{ Our Defense Models } To thoroughly compare the effectiveness of the random-patch based defense strategy, we perform the comparisons among the following three models as: (1) $\boldsymbol{Basic}$ is a standard defense model, in which a whole image with size $224\times224$ is as input for training. (2) $\boldsymbol{OurEven}$ is our defense model trained by dividing a whole image into nine patches with the same size and then resize every patch to $224\times224$. (3) $\boldsymbol{OurRandom}$ is our defense model trained by dividing a whole image into nine random patches and then resize every patch to $224\times224$.

\subsubsection{Backbones of Our Defense Models } To compare the performance of our defense strategy implemented by different backbone networks of the DNN, four different network structures are used as:
ResNet-18\cite{he2016deep} ("r" is for short.), Inception-v3\cite{szegedy2016rethinking}("i" is for short.), MobileNet-v3-small\cite{howard2019searching} ("m" is for short.) and VGG-11\cite{simonyan2014very} ("v" is for short.). Hence, each backbone network will be used in $Basic$, $OurEven$ and $OurRandom$, respectively. Thus we have $Basic_r$, $OurEven_r$ and $OurRandom_r$ which are three models implemented with ResNet-18. Similar settings are used for the other three backbone networks.

\begin{table}
\begin{center}
\caption{Parameters of the testing set generated by our multi-task attack model with a defense model and a face model. The specific defense model is listed in the "Defense-Model" column, and the face model is SE-LResNet101E-IR Arcface model. $\alpha$ is set empirically. $\beta$ is set randomly within a range.
}
\label{tab:testdata}
\setlength{\tabcolsep}{2mm}
\begin{tabular}{llccc}  
\hline\noalign{\smallskip}
  Dataset & Defense Model & $\alpha$ & $\beta$ & Size \\
\noalign{\smallskip} \hline \noalign{\smallskip}
$Test$    & -  & -  & - &1052    \\
$Test_{Basic}$    & $Basic_r$  & $1e^{-3}$  & $6e^{-5}-1e^{-4}$ &1052    \\
$Test_{Even}$    & $OurEven_r$   & $1e^{-3}$   & $6e^{-5}-1e^{-4}$     &1052\\
$Test_{Random}$    & $OurRandom_r$   & $1e^{-3}$  & $6e^{-5}-1e^{-4}$ &1052   \\
\hline
\end{tabular}
\end{center}
\end{table}

\subsubsection{Details of Various Attacks } To validate that our proposed defense method can successfully resist various attacks, including the ones which can attack the defense model, we generate two categories of attacks: white-box attacks and adaptive attacks, to verify the proposed defense models. 

\textbf{white-box Attack.} With two physical attacks (i.e., the hat-attack and the glasses-attack) implemented with either the dodging attack or the impersonation attack, we generate four white-box attacks as negative examples for every identity. All in together we have $526*2$ for each kind of physical attacks, and we label them as \textbf{Test} in the second row of Table \ref{tab:testdata}. These data will be use to train and verify our defense method (mentioned that all defense models are trained by same original images and white-box attacks).

\textbf{Adaptive Attack.} By changing either the defense strategy or the network of the defense model used in the multi-task attack model, we can generate different adaptive attacks. Without loss the generality, we train three defense models (i.e. $Basic_r$, $OurEven_r$, and $OurRandom_r$ ) by original training images and white-box attacks. Then, with the trained defense model and 526 original testing image, we use our proposed multi-task attack model to generate adaptive attacks. For each defense model, we generate 1052 hat-attacks (526 by the dodging attack and 526 by the impersonation attack) and 1052 glasses-attacks (526 by the dodging attack and 526 by the impersonation attack). For example, for the defense model $OurEven_r$, we obtain the adaptive attacks labeled with $Test_{Even}$, and the attack from $Test_{Even}$ is impressed with evenly splitting image patch as the defense strategy and ResNet-18 as the backbone network of the defense model. Hence, if $Test_{Even}$ is used to attack our defense model $OurEven_i$, it belongs to the defense-strategy-leaked attack. Meanwhile, if $Test_{Even}$ is used to attack our defense model $OurEven_r$, it belongs to the defense-model-leaked attack. Similar labels are used for other adaptive attacks. Later the generated testing data will be used to test the robustness of our proposed defense method. The detailed parameters to generate these adaptive attacks are shown in Table \ref{tab:testdata}.

\subsubsection{Evaluation Metrics } For evaluation metrics, we employ True Positive Rate (TPR) for the accurately detected normal images, and False Acceptance Rate (FAR) to measure the accuracy of the detected attacks. Hence, $FAR_{hat}$ is false acceptance rate for hat-attack, and $FAR_{glasses}$ is false acceptance rate for glasses-attack. The larger value of TPR the better the defense model is. For FAR, the smaller value the better the defense model is.

\subsection{The Effectiveness of the Multi-task Attack Model}
To validate the effectiveness of adaptive attacks generate by the multi-task attack model, we provide the performance comparisons under various settings in Table \ref{tab:sim}. Given a defense model or a face model, the effectiveness of an attack is measured by calculating the similarity of the attack to the target image (or the original image). For dodging attack, the large similarity the strong attack is, and it is vice verse for the impersonation attack. In this paper, we adopt the mean of cosine similarity. It is worth mentioning that in the row of \textbf{Original} in Table \ref{tab:sim} the original images are used as the attacks and act as the baseline of the performance. From Table \ref{tab:sim}, we can also see that under different settings, the similarity computed is totally different from the baseline especially the white-box attacks in row two (i.e. $Test$) and the adaptive attacks from row three to row five. To some extent, it manifests that it is essential to detect physical attacks in face recognition system. Meanwhile, for various attacks, even though some of them are particularly generated to attack a defense model (e.g., $Test_{Basic}$, $Test_{Even}$ and $Test_{Random}$), they almost have the similar strength to attack the Arcface model. In other words, just using a standard DNN (i.e. Arcface model) to detect attacks is not secure enough. It is especially true when the defense strategies are leaked, which occasionally happens in the real world.

\begin{table}
\begin{center}
\caption{Averaged cosine similarity between the attack and the target image. The row 'Original' means initial cosine similarity between each pair, and it is also worth mentioning that all test sets are used exactly same pairs to generate attacks. 
}
\label{tab:sim}
\setlength{\tabcolsep}{2mm}
\begin{tabular}{lcccccc}  
\hline\noalign{\smallskip}
 \multirow{3}*{$Dataset$} &
 \multicolumn{2}{c}{Hat-attack} & \multicolumn{2}{c}{Glasses-attack} \\
 \cmidrule(r){2-3}\cmidrule(r){4-5}
  & Dodging & Impersonation & Dodging & Impersonation
  \\
\cmidrule(r){1-1}\cmidrule(r){2-3}\cmidrule(r){4-5}
$Original$ & 1 & 0.0296 & 1 & 0.0296 \\
$Test$ & -0.0549 & 0.4573 & 0.2028 & 0.3054 \\
$Test_{Basic}$ & -0.0485 & 0.4479 & 0.2108 & 0.3062 \\
$Test_{Even}$ & -0.0495 & 0.4458& 0.2148 & 0.3054\\
$Test_{Random}$ & -0.0486 & 0.4471 & 0.2177 & 0.3033\\
\hline
\end{tabular}
\end{center}
\end{table}
\begin{table}
\begin{center}
\caption{Performance of our defense model against white-box attack. The "Model" column is the defense model to be evaluated.}
\label{tab:white-box}
\setlength{\tabcolsep}{3mm}
\begin{tabular}{lccc}  
\hline\noalign{\smallskip}
  Model & TPR & $FAR_{hat}$ & $FAR_{glasses}$ \\
 \noalign{\smallskip} \hline \noalign{\smallskip}
 $Basic_m$  & 100.00 & 0.00 & 0.00 \\
 $Basic_r$  & 99.81 & 0.00 & 0.00 \\
 $Basic_v$  & 99.81 & 0.19 & 0.00  \\
\noalign{\smallskip} \hline \noalign{\smallskip}
 $OurEven_m$  & 99.62 & 0.00 & 0.00  \\
 $OurEven_r$  & 100.00 & 0.00 & 0.00 \\
 $OurEven_v$  & 99.81 & 0.00 & 0.00  \\
 \noalign{\smallskip} \hline \noalign{\smallskip}
 $OurRandom_m$  & 99.43 & 0.00 & 0.00 \\
 $OurRandom_r$  & 100.00 & 0.00 & 0.00 \\
 $OurRandom_v$  & 100.00 & 0.00 & 0.00 \\
\noalign{\smallskip}
\hline
\end{tabular}
\end{center}
\end{table}
\subsection{Robustness to White-Box Attack}
To validate that our proposed defense model can resist white-box attack, followed by the definition of white-box attack in previous works \cite{tao2018attacks,wu2020defending}, we generate attacks which only have access to the face model, and test our methods under the generated white-box attacks. The detailed information of the generated white-box attack data can be referred to the row of \textbf{Test} in Table \ref{tab:testdata}. Without loss of generality, we take various defense models, and the results in Table \ref{tab:white-box} show that all defense models achieve good performance with large $TPR$ and small $FAR$ values. In other words, it's relatively simple to defend against attacks which have no access to the defense model or the defense strategies.

\begin{table}[!t]
\begin{center}
\caption{Performance of our defense model against the defense-strategy-leaked attacks. The "Dataset" column is the testing data, and the "Model" column is the defense model to be evaluated.}
\label{tab:dsl-aa}
\setlength{\tabcolsep}{2mm}
\begin{tabular}{lllll}  
\hline\noalign{\smallskip}
 Dataset & Model & TPR & $FAR_{hat}$ & $FAR_{glasses}$  \\
\noalign{\smallskip} 
\hline 
\noalign{\smallskip}
 & $Basic_{m}$    & 100  & 8.65  & 5.89 \\
$Test_{Basic}$ & $Basic_i$    & 99.81  & 16.06  & 3.30 \\
 & $Basic_v$    & 99.81  & 46.58  & 18.54 \\ \noalign{\smallskip} \hline \noalign{\smallskip}
 & $OurEven_m$    & 99.62  & 0.00  & 0.00 \\
$Test_{Even}$ & $OurEven_{i}$    & 100  & 0.00  & 0.10 \\
 & $OurEven_{v}$    & 99.81  & 0.00  & 0.10 \\ \noalign{\smallskip} \hline \noalign{\smallskip}
 & $OurRandom_{m}$    & 98.67  & 0.00  & 0.57 \\
$Test_{Random}$ & $OurRandom_{i}$    & 100  & 0.00  & 0.00 \\
 & $OurRandom_{v}$    & 100  & 3.80  & 0.10 \\
\hline
\end{tabular}
\end{center}
\end{table}

\begin{table}[t]
\begin{center}
\caption{Performance  of our defense model against the defense-model-leaked attack. The "Dataset" column is the testing data, and the "Model" column is the defense model to be evaluated.}
\label{tab:dml-aa}
\setlength{\tabcolsep}{2mm}
\begin{tabular}{lllll}  
\hline\noalign{\smallskip}
  Dataset & Model & TPR & $FAR_{hat}$ & $FAR_{glasses}$  \\
 \noalign{\smallskip}\hline\noalign{\smallskip}
$Test_{Basic}$ &  $Basic_r$    & 99.81  & 100  & 99.14 \\
$Test_{Even}$ &  $OurEven_r$    & 100  & 95.25  & 68.63 \\
$Test_{Random}$ & $OurRandom_r$    & 100  & 59.13  & 36.50 \\
\hline
\end{tabular}
\end{center}
\end{table}

\subsection{Robustness to Defense-Strategy-Leaked Attack}
In order to further validate the robustness of our proposed defense model, we conduct the performance comparisons for the defense-strategy-leaked attacks. For example, the testing set $Test_{Basic}$ is generated by the multi-task attack model with the basic defense model (i.e., ResNet-18 and detection on the whole image) and the SE-LResNet101E-IR Arcface model. Since it is the defense strategy leaked attack, we can use it to attack a defense model with the same defense strategy but a different backbone network, which could be $Basic$ with other DNN models (i.e., $Basic_m$, $Basic_i$ and $Basic_v$). In this way, we can verify our proposed defense model via the defense-strategy-leaked attacks. The result is shown in Table \ref{tab:dsl-aa}. We can find that the standard defense method $Basic$ even with different DNN backbone models is easily attacked with FAR ranging from $3.30$ to $46.58$. On the contrary, our proposed defense methods $OurEven$ and $OurRandom$ both work well against the defense-strategy-leaked attack with FAR ranging from $0$ to $3.8$. These further validate that our patch based defense strategies are effective.

One special phenomenon is that VGG model is more vulnerable to be attacked, such as $Basic_{v}$ on the fourth row, $OurEven_{v}$ on the seventh row and $OurRandom_{v}$ on the tenth row all have relative larger $FAR_{hat}$ and $FAR_{glasses}$ than the corresponding defense methods with other backbone networks. It is generally because VGG is too large for this detection task and overfitted the distribution of the training data. The different distribution caused by different attack processes makes VGG models perform poorer than other small network models (e.g., ResNet-18, Inception-v3, MobileNet-v3-small). We also find $OurRandom$ performs worse than $OurEven$ the defense-strategy-leaked attack. This is because attacks in $Test_{Random}$ are stronger than $Test_{Even}$.  We will show $OurRandom$ is more robust than $OurEven$ in the following subsection against the defense-model-leaked attacks. 

\subsection{Robustness to Defense-Model-Leaked Attack}
In the defense-model-leaked attacks, the attacker knows both the model parameters and the defense strategy. For example, the testing set $Test_{Random}$ is generated by the multi-task attack model, which attacks the standard DNN (i.e.ResNet-18 as the backbone and the randomly split image patch as the defense strategy) and the SE-LResNet101E-IR Arcface model. Hence, we use it to attack the defense model with the same parameters and the same defense strategy, which should be $OurRandom_r$. In this way, we can verify our proposed defense model via the defense-model-leaked attacks. We can see in Table \ref{tab:dml-aa} that the basic defense model $Basic_r$ almost can't resist any attack with the defense-model-leaked attack settings (i.e. 100 of $FAR_{hat}$ and 99.14 of $FAR_{glasses}$). On the contrary, our proposed defense model $OurEven_r$ is a little better, and $OurRandom_r$ detects about 40\% attacks in the hat-attack and more than 60\% attacks in the glasses-attack. The robustness of our $OurRandom_r$ defense model to the defense-model-leaked attack benefits by the randomness of image patches in the defense strategy.

\begin{table*}[!h]
\begin{center}
\caption{The detection performance of our defense methods with  attacks only attacking different Arcface models. }
\label{tab:detectinfo}
\begin{tabular}{p{2cm}p{0.8cm}cccccc} 
\hline \noalign{\smallskip}
& & \multicolumn{2}{c}{\footnotesize{SE-LResNet101E-IR}} & \multicolumn{2}{c}{SE-LResNet50E-IR} & \multicolumn{2}{c}{MobileFaceNet} \\
\cmidrule(r){3-4} \cmidrule(r){5-6} \cmidrule(r){7-8} 
Model & TPR & $FAR_{hat}$ & $FAR_{glasses}$ & $FAR_{hat}$ & $FAR_{glasses}$ & $FAR_{hat}$ & $FAR_{glasses}$  \\
\cmidrule(r){1-2} \cmidrule(r){3-4} \cmidrule(r){5-6} \cmidrule(r){7-8} 
 $Basic_r$  & 99.81 & 0.00 & 0.00 & 3.14 & 0.00 & 16.16 & 0.00  \\
 $OurEven_r$  & 100 & 0.00 & 0.00 & 0.00 & 0.00 & 0.00 & 0.00 \\
 $OurRandom_r$  & 100 & 0.00 & 0.00 & 0.00 & 0.00 & 0.00 & 0.00 \\
\hline
\end{tabular}
\end{center}
\end{table*}

\begin{figure}
    \begin{minipage}[t]{0.48\linewidth}
        \centering
        \includegraphics[width=1.5in]{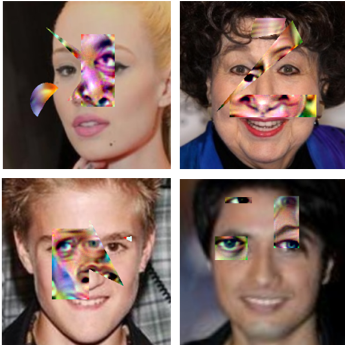}
    \end{minipage}
    \begin{minipage}[t]{0.48\linewidth}
        \centering
        \includegraphics[width=1.5in]{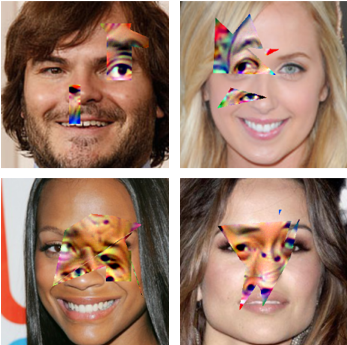}
    \end{minipage}
    \caption{Illustration of attacks with random masks. Left is faces from celebA, and right is faces from VGGface.}
    \label{fig.random-mask}
\end{figure}

\begin{table}
\begin{center}
\caption{Cross verification of our proposed defense model on celebA and VGGface datsets.}
\label{cross-verify}
\setlength{\tabcolsep}{6mm}
\begin{tabular}{llcc}  
\hline\noalign{\smallskip}
  Training  & Testing  & TPR & FAR \\
 \noalign{\smallskip}\hline\noalign{\smallskip}
VGGface &  celebA   & 99.97  & 0 \\
celebA &  VGGface   & 96.25  & 0 \\
\hline
\end{tabular}
\end{center}
\end{table}

\begin{table}
\begin{center}
\caption{Performance comparisons of our defense model to attacks with random masks on two datasets: VGGface and celebA.}
\label{random-mask}
\setlength{\tabcolsep}{8mm}
\begin{tabular}{lcc}  
\hline\noalign{\smallskip}
  Dataset & TPR & FAR \\
 \noalign{\smallskip}\hline\noalign{\smallskip}
VGGface   & 95.63  & 0 \\
celebA   & 99.80  & 0 \\
\hline
\end{tabular}
\end{center}
\end{table}

\begin{table}[t]
\setlength{\tabcolsep}{3mm}
\begin{center}
\caption{By varying the values of the parameter $\alpha$ in our multi-task attack model to generate new testing data (i.e. adaptive attacks), we compare the detection performance between $Basic_m$ and $OurRandom_m$. }
\label{tab:attack-compare}
\begin{tabular}{lllll} 
\hline \noalign{\smallskip} 
\multirow{2}*{$\alpha$} &
\multicolumn{2}{c}{$Basic_m$} & \multicolumn{2}{c}{$OurRandom_m$} \\
\cmidrule(r){2-3} \cmidrule(r){4-5}
 & $FAR_{hat}$ & $FAR_{glasses}$ &  $FAR_{hat}$ & $FAR_{glasses}$ \\
 0.001  & 8.65 & 5.89  & 0.00 & 0.57 \\
 0.01  & 18.54 & 22.15 & 0.48 & 3.42 \\
 0.1  & 22.53 & 20.34 & 3.14 & 9.13 \\
\hline
\end{tabular}
\end{center}
\end{table}

\subsection{Ablation Study}

\subsubsection{Generalization to Different Datasets } \label{subsection:dataset}
In this part, we use two different datasets to cross verify our methods. We pick one image from every identity in celebA\cite{liu2015faceattributes} and VGGface datasets, and then get 10177 images of celebA and 2622 images of VGGface datasets. For every image, by attacking SE-LResNet101E-IR Arcface model we generate one dodging hat-attack and one dodging glasses-attack. Finally, we perform cross verification on 20354 attacks of celebA and 5244 attacks of VGGface dataset for our proposed defense model $OurRandom_m$. Specifically, if $OurRandom_m$ is trained with 5244 attacks of VGGface dataset, then it is tested by 20354 attacks of celebA, and vice versa. The results in Table \ref{cross-verify} show that our defense method $OurRandom_m$ is robust on different datasets.

\subsubsection{Robustness to Different Shapes of Attacks }
Considering that the attacker may generate different shapes of attacks, in order to show that our proposed defense model is robust to the attacks with various shapes, we design an experiment in which we are not available to the shape of attacks, and our model is trained by attacks with random masks. A random mask is generated by stacking several random polygons such as triangle, rectangle and circle, etc.. Some examples of attacks generated with random masks are shown in Fig.\ref{fig.random-mask}. Then, as two special cases of attacks with random masks, the hat-attack and the glass-attack are used to test the trained model. We use the same dataset in the above subsection but split each dataset into 7:2:1 as the training set, the validation set and the testing set. By attacking SE-LResNet101E-IR Arcface model, we obtain attacks with random masks for training, and attacks with forehead region (i.e. hat-attack) and glasses region (i.e. glass-attack) for testing. Still, we use the model $OurRandom_m$ as the defense model, and the detection results shown in Table \ref{random-mask} further validate that our proposed defense model is robust to different shapes of attacks.

\subsubsection{Robustness to Various Face Models}
To show our proposed defense method is robust to various face models, we compare the detection accuracy on three datasets, which are generated by attacking three different backbones of the Arcface model: SE-LResNet101E-IR, SE-LResNet50E-IR, and MobileFaceNet. The detailed results are shown in Table \ref{tab:detectinfo}. $Basic_r$, $OurEven_r$ and $OurRandom_r$ are used as the target defense models for the evaluation of this part.

As shown in Table \ref{tab:detectinfo}, the standard DNN ($Basic_r$) achieves 100\% accuracy (i.e. 0.00 of both $FAR_{hat}$ and $FAR_{glasses}$) against attacks when the training data and the testing data are both generated only by attacking SE-LResNet101E-IR Arcface model. But the accuracy rate is decreased when the Arcface model changes its backbones (e.g., 16.16 of $FAR_{hat}$ for MobileFaceNet). While the proposed $OurEven_r$ and $OurRandom_r$ both show the stable performance on $FAR$, and achieve 100\% $TPR$, which is slightly higher than the standard DNN.

\subsubsection{Parameter Analysis}
Since $\alpha$ in Eq.(1) is an important parameter to our proposed defense model, to validate its sensitivity, we conduct the experiments by adjusting its values in our multi-task attack model to generate new testing data. Then using these data to attack our defense model, we check whether the performance varies as $\alpha$ changing. In other words, with different $\alpha$, we obtain different $Test_{Random}$, then we use these data to attack $OurRandom_{m}$ and $Basic_m$, in which by using the defense-strategy-leaked attack we validate the two most efficient defense models.

Specifically, by increasing the values of $\alpha$ from 0.001 to 0.1 gradually, we can generate new attacks as the testing data by attacking the SE-LResNet101E-IR Arcface model and the detection model either $Basic_r$ or $OurRandom_r$. Then with these generated testing data, we compare the performance of  $OurRandom_m$ with $Basic_m$, regarding to $FAR_{hat}$ and $FAR_{glasses}$ as shown in Table \ref{tab:attack-compare}. From this table, we can see that $OurRandom_m$ is still efficient with low $FAR$ when $\alpha$ is increasing, while the performance of $Basic_m$ dropped sharply. These experimental results show that our method is much more robust than standard DNN, and still robust in a certain range of $\alpha$. In other words, $\alpha$ is not sensitive to our defense model. The increment of FAR when $\alpha=0.1$ is due to the gap between the testing attacks and the training data. Hence, data augmentation, such as color conversion and adding noises, can improve the performance of detection models. Also, the most efficient way is adding data generated by the new attacks, which is similar to adversarial training, while this part is out the range of this paper, and we make it as one of our future work.

\section{Conclusion}
In this paper, we propose a random-patch based defense method to detect physical attacks on face recognition system in the real world. Although the defense strategy by randomly splitting the image into patches is straightforward even simple, the excellent detection performance against the white-box attacks and adaptive attacks validates its effectiveness and robustness. Moreover, extensive experimental results on VGGface and celebA datasets further show the superiority of the proposed defense method to resist attacks with random masks even on different datasets. Additionally, our proposed method can conveniently combine with existing defense method(such as adversarial training and feature squeezing) to further boost the performance. 



\section*{Acknowledgment}
The authors would like to thank anonymous reviewers for their valuable suggestions. This work was supported by the General Program of National Natural Science Foundation of China (NSFC) (Grant No. 61806147).
This research was also partially supported by Shanghai Natural Science Foundation of China (Grant No. 18ZR1441200).

\ifCLASSOPTIONcaptionsoff
  \newpage
\fi



%
\bibliographystyle{IEEEtran}
\bibliography{egbib}

%







\end{document}